\documentclass{article}

\PassOptionsToPackage{numbers, compress}{natbib}

\usepackage[final]{nips_2018}      % [][preprint][final]

\usepackage[utf8]{inputenc} % allow utf-8 input
\usepackage[T1]{fontenc}    % use 8-bit T1 fonts
\usepackage{hyperref}       % hyperlinks
\usepackage{url}            % simple URL typesetting
\usepackage{booktabs}       % professional-quality tables
\usepackage{amsfonts}       % blackboard math symbols
\usepackage{nicefrac}       % compact symbols for 1/2, etc.
\usepackage{microtype}      % microtypography
\usepackage{amsmath}        % ... for \text{}
\usepackage{subfig}
\usepackage{bm}             % ... for \bm{}
\usepackage{graphicx}       % ... for \includegraphics
\usepackage{float}          % ... for [H]
\usepackage{makecell}       % ... for \thead and \\
\usepackage{xcolor}         % ... for \textcolor
\usepackage{pifont}
\newcommand{\cmark}{\ding{51}}
\newcommand{\xmark}{\ding{55}}

\graphicspath{{images/}}

\title{\textsc{Match}-Net: Dynamic Prediction in Survival Analysis using Convolutional Neural Networks}

\author{
Daniel Jarrett$^{*}$, Jinsung Yoon$^{\dagger}$, and Mihaela van der Schaar$^{*,\dagger}$ \\
$^{*}$ Department of Engineering Science, University of Oxford, UK \\
$^{\dagger}$ Department of Electrical Engineering, University of California Los Angeles, USA \\
\texttt{daniel.jarrett@eng.ox.ac.uk}, \texttt{jsyoon0823@ucla.edu}, \texttt{mihaela@ee.ucla.edu} \\
}

\begin{document}

\maketitle

\begin{abstract}
Accurate prediction of disease trajectories is critical for early identification and timely treatment of patients at risk. Conventional methods in survival analysis are often constrained by strong parametric assumptions and limited in their ability to learn from high-dimensional data, while existing neural network models are not readily-adapted to the longitudinal setting. This paper develops a novel convolutional approach that addresses these drawbacks. We present \textsc{Match}-Net: a Missingness-Aware Temporal Convolutional Hitting-time Network, designed to capture temporal dependencies and heterogeneous interactions in covariate trajectories and patterns of missingness. To the best of our knowledge, this is the first investigation of temporal convolutions in the context of dynamic prediction for personalized risk prognosis. Using real-world data from the Alzheimer's Disease Neuroimaging Initiative, we demonstrate state-of-the-art performance without making any assumptions regarding underlying longitudinal or time-to-event processes\textemdash attesting to the model's potential utility in clinical decision support.
\end{abstract}

\section{Introduction}

In Alzheimer's disease\textemdash the annual cost of which exceeds \$800 billion globally \cite{marinescu2018tadpole}\textemdash the effectiveness of therapeutic treatments is often limited by the challenge of identifying patients at early enough stages of  progression for treatments to be of use. As a result, accurate and personalized  prognosis in earlier stages of cognitive decline is critical for effective intervention and subject selection in clinical trials. Conventional statistical methods in survival analysis often begin by choosing explicit functions to model the underlying stochastic process \cite{doksum1992models, kleinbaum2005survival, rodriguez2005parametric, lee2010proportional, fernandez2016gaussian, alaa2017deep}. However, the constraints\textemdash such as linearity and proportionality \cite{singh2011survival, cox1992regression} in the popular Cox model\textemdash may not be valid or verifiable in practice.

In spite of active research, a conclusive understanding of Alzheimer's disease progression remains elusive, owing to heterogeneous biological pathways \cite{beckett2015alzheimer, hardy2002amyloid}, complex temporal patterns \cite{jedynak2012computational, donohue2014estimating}, and diverse interactions \cite{frolich2017incremental, pascoal2017synergistic}. Hence Alzheimer's data is a prime venue for leveraging the potential advantages of deep learning models for survival. Neural networks offer versatile alternatives by virtue of their capacity as general-purpose function approximators, able to learn\textemdash without restrictive assumptions\textemdash the latent structure between an individual's prognostic factors and odds of survival.

\textbf{Contributions}. Our goal is to establish a novel convolutional model for survival prediction, using Alzheimer's disease as a case study for experimental validation. Primary contributions are threefold: First, we formulate a \textit{generalized framework} for longitudinal survival prediction, laying the foundation for effective cross-model comparison. Second, our proposal is uniquely designed to capitalize on longitudinal data to issue \textit{dynamically updated} survival predictions, accommodating potentially informative patterns of \textit{data missingness}, and combining \textit{human input} with model predictions. Third, we propose methods for deriving clinically meaningful insight into the model's inference process. Finally, we demonstrate state-of-the-art results in comparison with comprehensive benchmarks.

\section{Problem formulation}

Let there be $N$ patients in a study, indexed $i\in\{1,...,N\}$. Time is treated as a discrete dimension of fixed resolution $\delta>0$. Each longitudinal datum consists of the tuple $(t,\mathbf{x}_{i,t},s_{i,t})$, where $\mathbf{x}_{i,t}$ is the vector of  covariates recorded at time $t$, and $s_{i,t}$ is the binary survival indicator. Let random variable $T_{i,\text{surv}}$ denote the time-to-event, $T_{i,\text{cens}}$ the time of right-censoring, and $T_{i}=\min\{T_{i,\text{surv}}, T_{i,\text{cens}}\}$. Per convention, we assume that censoring is not correlated with the eventual outcome \cite{tsiatis2004joint, zheng2005partly, van2011dynamic}. Let the complete longitudinal dataset be given by $\mathbf{X}=\{\langle(t,\mathbf{x}_{i,t},s_{i,t})\rangle^{t_i}_{t=0}\}^N_{i=1}$. Then we can define\begin{equation}
\mathbf{X}_{i,t,w}=\langle(t^{\prime},\mathbf{x}_{i,t^{\prime}},s_{i,t^{\prime}})\rangle_{t^{\prime}\in T}
\text{\space\space\space where\space\space\space}
T=\{t^{\prime}:t-w\leq t^{\prime}\leq t\}
\end{equation}
to be the set of observations for patient $i$ extending from time $t$ into a width-$w$ window of the past, where parameter $w$ depends on the model under consideration.  Given longitudinal measurements in $\mathbf{X}_{i,t,w}$, our goal is to issue risk predictions corresponding to length-$\tau$ horizons into the future. Formally, given a \textit{backward-looking} historical window $(t-w,t]$, we are interested in the failure function for \textit{forward-looking} prediction intervals $(t,t+\tau]$; that is, we want to estimate the probability\begin{equation}
F_i(t+\tau|t,w)=\mathbb{P}(T_{i,\text{surv}}\leq t+\tau|T_{i,\text{surv}}>t,\mathbf{X}_{i,t,w})
\end{equation}
of event occurrence within each prediction interval. Observe that parameterizing the width of the historical window results in a generalized framework\textemdash for instance, a Cox model only utilizes the most recent measurement; that is, $w=1$. At the other extreme, recurrent models may consume the entire history; that is, $w=t$. As we shall see, the best performance is in fact obtained via a flexible intermediate approach\textemdash that is, by learning the optimal width of a sliding window of history.

\section{MATCH-Net}

\begin{figure}[H]
\centering
\includegraphics[width=\linewidth]{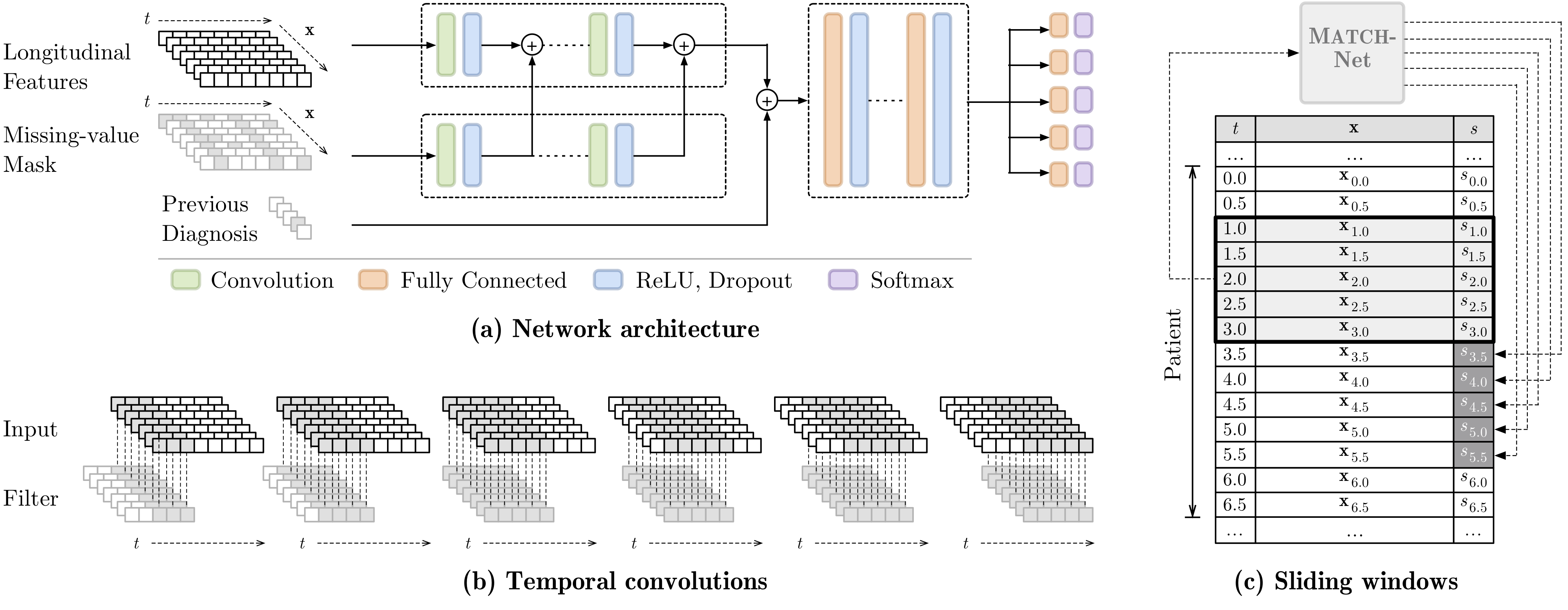}
\caption{(a) The \textsc{Match}-Net architecture, with $\tau_\text{max}=5\delta$. (b) Illustration of temporal convolutions acting over feature channels. (c) The longitudinal context within which \textsc{Match}-Net operates, as well as the network's prediction targets in association with the sliding window input mechanism.}
\end{figure}

We propose \textsc{Match}-Net: a Missingness-Aware Temporal Convolutional Hitting-time Network, innovating on current approaches in three respects. \textbf{Dynamic prediction}: Existing deep learning models issue prognoses on the basis of information from a single time point \cite{faraggi1995neural, buckley1979linear, katzman2016deep, xiang2000comparison, mariani1997prognostic, liestbl1994survival, biganzoli2009partial, yu2011learning, luck2017deep, fotso2018deep, lee2018deephit}, potentially discarding valuable information in the presence of longitudinal data. We investigate \textit{temporal convolutions} in capturing heterogeneous representations of temporal dependencies among observed covariates, enabling truly dynamic predictions on the basis of historical information. \textbf{Informative missingness}: Current survival methods rely on the common assumption that the timing and frequency of covariate measurements is uninformative \cite{van2011dynamic, rizopoulos2012joint}. By contrast, our model is \textit{missingness-aware}\textemdash that is, we explicitly account for informative missingness by learning correlations between patterns of missingness and disease progression. \textbf{Human input}: Instead of issuing predictions solely on the basis of quantitative clinical measurements, we optionally incorporate clinicians' most recent diagnoses of patient state into model estimates to examine the incremental informativeness of \textit{subjective} input. Each innovation is a source of gain in performance (see Section \ref{experiments}; further details in the appendix).

\textsc{Match}-Net accepts as input a \textit{sliding-window} of observed covariates $\mathbf{X}_{i,t,w}$, as well as a parallel binary mask of missing-value indicators $\mathbf{Z}_{i,t,w}$ taking on values of one to denote missing covariate measurements. In addition, the network optionally accepts a one-hot vector $\mathbf{r}_{i,t}$ describing the most recent clinician diagnosis of Alzheimer's disease progression. Starting from the base of the network, the convolutional block first learns representations of longitudinal covariate trajectories by extracting local features from temporal patterns. After each layer, filter activations from the auxiliary branch are concatenated with those in the main branch. Then, the fully-connected block captures more global relationships by combining local information. Finally, the output layers produce failure estimates \begin{equation}
\hat{\mathbf{y}}_{i,t}=[\hat{F}_i(t+\delta|t,w),...,\hat{F}_i(t+\tau_\text{max}|t,w)]
\end{equation}
for pre-specified prediction intervals, where $\tau_\text{max}$ is the maximal horizon desired. This \textit{convolutional dual-stream} architecture explicitly captures representations of temporal dependencies within each stream, as well as between covariate trajectories and missingness patterns in association with disease progression. This accounts for the potential informativeness of both \textit{irregular} sampling (\textit{i.e.} intervals between consecutive clinical visits may vary) and \textit{asynchronous} sampling (\textit{i.e.} not all features are measured at the same time) \cite{rubin1976inference, bangphased}. This also encourages the network to distinguish between actual measurements and imputed values, reducing sensitivity to specific imputation methods chosen. The final architecture uses convolutions of length $l=5\delta$, with more filters per layer in the main branch.

\textbf{Loss function}. With the preceding notation, the negative log-likelihood of a single empirical result $s_{i,t+\tau}$ and model estimate $\hat{F}_i(t+\tau|t,w)$ in relation to some input window $\mathbf{X}_{i,t,w}$ is given by
\begin{equation}
l_{i,t,\tau}(\bm{\theta})=-[s_{i,t+\tau}\log \hat{F}_i(t+\tau|t,w)+(1-s_{i,t+\tau})\log (1-\hat{F}_i(t+\tau|t,w))]
\end{equation}
where $\bm{\theta}$ denotes the parameters of the network. The total loss function is then computed to simultaneously take into account the quality of predictions for all prediction horizons $\tau$, all times $t$ available along each patient's longitudinal trajectory, and all patients $i\in\{1,...,N\}$ in the survival dataset:
\begin{equation}
l(\bm{\theta})=\frac{\delta^2}{\sum_{i=1}^N\sum_{j=1}^{t_i} \tau_i}\text{ }\sum_{i=1}^N\text{ }\sum_{j=1}^{t_i/\delta}\text{ }\sum_{k=1}^{\tau_i/\delta} \alpha(i,(j-1)\delta,k\delta)\cdot l_{i,(j-1)\delta,k\delta}
\end{equation}
where $\tau_{i}=\min\{t_i-t,\tau_\text{max}\}$ accounts for failure or right-censoring. This is a natural generalization of the log-likelihood in \cite{liestbl1994survival} to accommodate longitudinal survival. Weight function $\alpha(i,t,\tau)$ allows trading off the relative importance of different patients, time steps, and prediction horizons. First, this allows standardizing patient contributions with $\alpha(i,t,\tau)\propto 1/t_i$, thereby counteracting the bias against patients with shorter survival durations. Second, in the context of heavily imbalanced classes, this allows up-weighting positive instances\textemdash that is, input windows that correspond to actual failure.

\section{Experiments}
\label{experiments}

The Alzheimer's Disease Neuroimaging Initiative (ADNI) study data is a longitudinal survival dataset of per-visit measurements for 1,737 patients \cite{marinescu2018tadpole}. The data tracks disease progression through clinical measurements at \nicefrac{1}{2}-year intervals, including quantitative biomarkers, cognitive tests, demographics, and risk factors. Our objective is to predict the \textit{first stable occurrence} of Alzheimer's disease for each patient. Further information on the dataset, preparation, and training can be found in the appendix.

\textbf{Benchmarks}. We evaluate \textsc{Match}-Net against both traditional longitudinal methods in survival analysis and recent deep learning approaches; the former includes Cox landmarking and joint modeling methods, and the latter includes static and dynamic multilayer perceptrons and recurrent neural network models. Performance is evaluated on the basis of the area under the receiver operating characteristic curve (AUROC) and the area under the precision-recall curve (AUPRC), both computed with respect to prediction horizons $\tau$. Five-fold cross validation metrics are reported in Table \ref{benchmark_results}.

\textbf{Performance}. \textsc{Match}-Net produces state-of-the-art results, consistently outperforming both conventional statistical and neural network benchmarks. Gains are especially apparent in AUPRC scores\textemdash improving on the MLP by an average of 15\% and on joint models by 16\% across all horizons, and by 27\% and 26\% for one-step-ahead predictions. To understand the sources of improvement, we observe a 4\% gain in AUPRC from introducing the sliding window mechanism (MLP to S-MLP), a 9\% gain from incorporating temporal convolutions (S-MLP to S-TCN), and a further 2\% gain from accommodating informative missingness (S-TCN to \textsc{Match}-Net). In addition, \textit{including} the most recent clinician diagnosis results in a further 17\% gain (further details located in the appendix).

\begin{table}
\caption{Cross validation performance for $\tau_\text{max}=5\delta$ and $\delta=$ \nicefrac{1}{2} years: \textsc{Match}-Net \textit{without} clinician input, sliding-window temporal convolutional networks (S-TCN) and multilayer perceptrons (S-MLP). Benchmarks include fully-convolutional networks (FCN) \cite{long2015fully} adapted for sequence-based survival prediction, Disease Atlas (D-Atlas) \cite{lim2018forecasting}, baseline recurrent neural networks (RNN) including GRUs and LSTMs, static multilayer perceptrons (MLP), as well as conventional statistical methods for survival analysis\textemdash including joint modeling (JM) and Cox landmarking (LM). Bold values indicate best performance, and asterisks next to benchmark results indicate statistically significant difference ($p$-value < 0.05) from \textsc{Match}-Net result. More detailed breakdown of gains are found the appendix.\medskip}
\label{benchmark_results}
\centering
\setlength\tabcolsep{3.6pt}
\begin{tabular}{c|c|ccc|cccccc} \toprule
{} & {$\tau$} & {\textsc{Match}-Net} & {S-TCN} & {S-MLP} & {FCN} & {D-Atlas} & {RNN} & {MLP} & {JM} & {LM} \\ \midrule
\textsc{auroc} & 0.5 & \textbf{0.962~} & 0.961~ & 0.959~ & 0.954~ & 0.959~ & 0.949* & 0.948* & 0.913* & 0.909* \\
               & 1.0 & \textbf{0.942~} & 0.941~ & 0.932~ & 0.930~ & 0.929~ & 0.930~ & 0.930~ & 0.917* & 0.914* \\
               & 1.5 & \textbf{0.902~} & \textbf{0.902~} & 0.897~ & 0.895~ & 0.892~ & 0.891~ & 0.890~ & 0.881~ & 0.878~ \\
               & 2.0 & \textbf{0.909~} & 0.908~ & 0.904~ & 0.903~ & 0.896~ & 0.901~ & 0.895~ & 0.894~ & 0.890~ \\
               & 2.5 & \textbf{0.886~} & 0.884~ & 0.881~ & 0.883~ & 0.884~ & 0.883~ & 0.874~ & 0.883~ & 0.878~ \\ \midrule
\textsc{auprc} & 0.5 & \textbf{0.594~} & 0.580~ & 0.500~ & 0.536~ & 0.517~ & 0.464* & 0.469* & 0.473* & 0.469* \\
               & 1.0 & \textbf{0.513~} & 0.505~ & 0.447~ & 0.453~ & 0.423~ & 0.410* & 0.435~ & 0.415* & 0.412* \\
               & 1.5 & \textbf{0.373~} & 0.367~ & 0.354~ & 0.357~ & 0.364~ & 0.340~ & 0.340~ & 0.319~ & 0.325~ \\
               & 2.0 & \textbf{0.390~} & 0.380~ & 0.364~ & 0.375~ & 0.352~ & 0.355~ & 0.359~ & 0.362~ & 0.367~ \\
               & 2.5 & \textbf{0.384~} & 0.381~ & 0.371~ & 0.365~ & 0.360~ & 0.365~ & 0.356~ & 0.366~ & 0.363~ \\
\bottomrule
\end{tabular}
\end{table}

\begin{figure}[H]
\centering
\vspace{-2mm} %%%
\includegraphics[width=\linewidth]{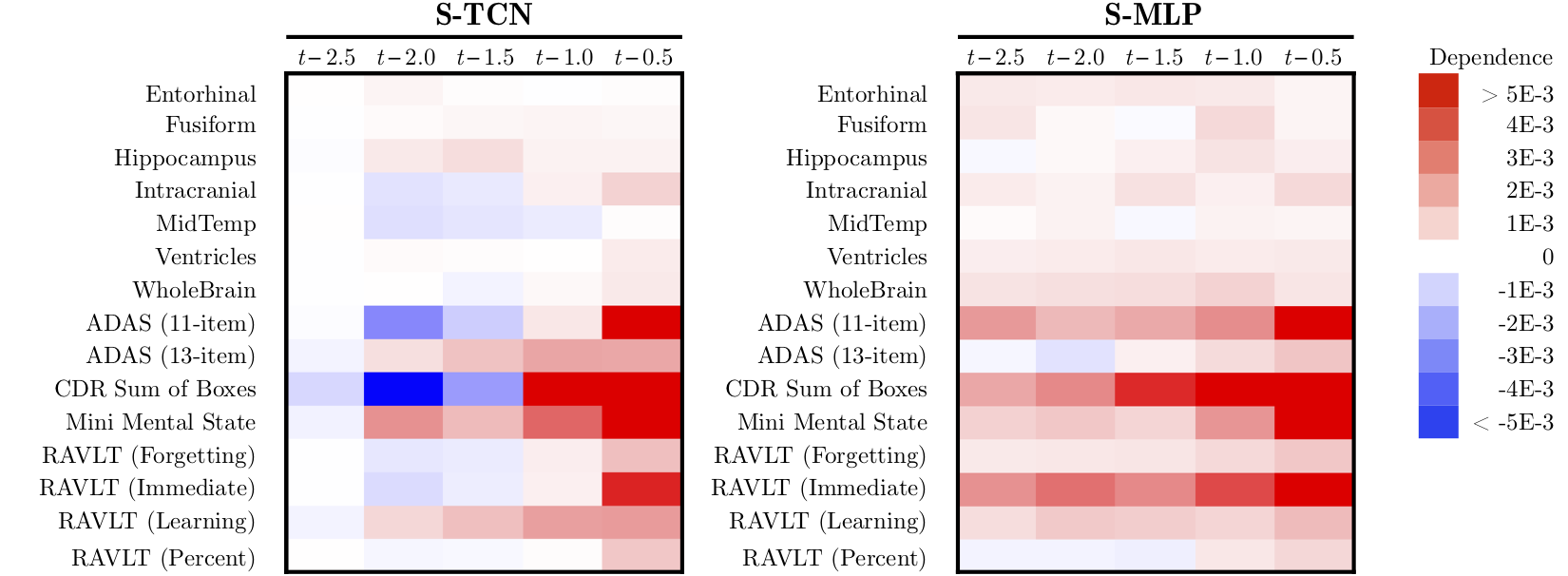}
\caption{Average saliency map indicating feature and temporal influence, computed using slopes of partial dependence on sample of numerical features across sliding window ($w=5\delta$; $\delta=$ \nicefrac{1}{2} years).}
\vspace{-2mm} %%%
\end{figure}

\textbf{Visualization}. From the preceding, we observe the largest gains by introducing convolutions. While this is consistent with our motivating hypothesis that convolutions are better able to capture temporal patterns, clinicians often desire a degree of transparency into the prediction process \cite{ribeiro2016should, avati2017improving}.  We adopt the partial dependence approach in \cite{friedman2001greedy} to understand the input-output relationship, as well as examining the utility of convolutions. For each observed covariate $d$, we want to approximate how the estimated failure function varies based on the value of $\mathbf{x}^d_{t,w}$. We define the dependence
\vspace{-1mm} %%%
\begin{equation}
\label{partial_dependence}
\mathbb{E}_{\mathbf{X}^{(-d)}_{t,w}}[\hat{F}(t+\tau|\mathbf{X}_{t,w})]\approx\frac{\delta}{\sum^N_{i=1}t_i}\sum_{i=1}^N\sum_{j=1}^{t_i/\delta}\hat{F}((j-1)\delta+\tau|\mathbf{x}^{d}_{(j-1)\delta,w},\mathbf{X}^{(-d)}_{(j-1)\delta,w})
\vspace{-1mm} %%%
\end{equation}
where $\mathbf{x}^d_{t,w}\cup\mathbf{X}^{(-d)}_{t,w}=\mathbf{X}_{t,w}$. By evaluating Equation \ref{partial_dependence} on the values $\mathbf{x}^d_{t,w}$ present in the data, the influence of each covariate can be measured by estimating its slope. For a global picture of what impact each feature and time step has on the model's predictions, we compute the influence for all features to produce an average \textit{saliency map} \cite{zeiler2014visualizing, simonyan2013deep} highlighting the effect of convolutional layers. All else equal, absent convolutions we see that having worse covariate values any time step almost invariably has an upward impact on risk (\textit{i.e.} negative impact on survival). On the other hand, with temporal convolutions we see that having worse covariate values at earlier time steps may result in a downward impact on risk (\textit{i.e.} positive impact on survival), suggesting that convolutions may better facilitate modeling relative movements (\textit{e.g.} sudden declines) than simply paying attention to levels. Further visualizations for added perspective on input-output relationships are found in the appendix.

\bibliographystyle{unsrtnat}
{\footnotesize
\bibliography{references}}

\clearpage

\section*{\centering\textsc{Appendix}}

\section*{Example use}

\begin{figure}[H]
\centering
\includegraphics[width=0.97\linewidth]{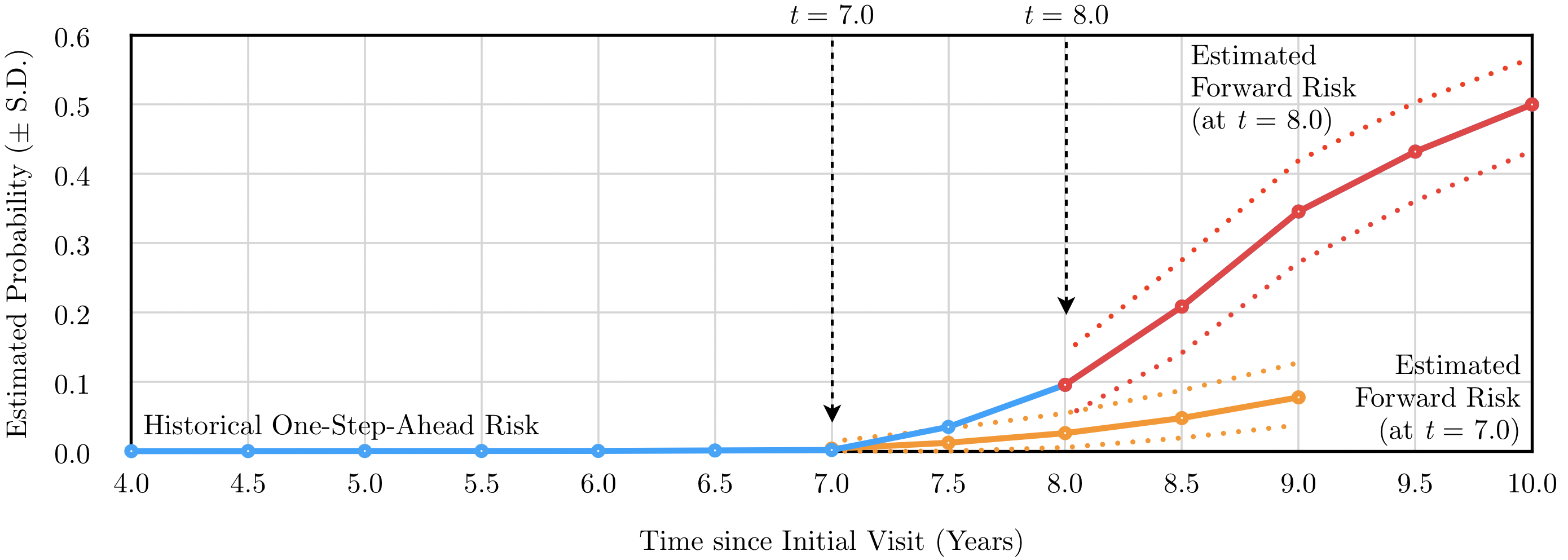}
\caption{Example application of \textsc{Match}-Net for personalized risk scoring.}
\label{use_case}
\end{figure}

While various settings may benefit from \textsc{Match}-Net as a matter of clinical decision support, we illustrate one possible application within the context of personalized screening. Figure \ref{use_case} shows the historical risk trajectory and forward risk estimates for a randomly selected ADNI patient. During the first seven years of bi-annual visits, the patient exhibits cognitively normal behavior, and the \textsc{Match}-Net risk estimates\textemdash computed via measured biomarkers and neuropsychological tests\textemdash reflect this steady clinical state (see historical trajectory in blue). In fact, as of precisely seven years of follow-up, the predicted 30-month forward risk remains less than 10\% (see predicted trajectory in orange).

However, two clinical visits later, when the patient returns for regular checkup and clinical measurements, the projected 30-month forward risk jumps to over 50\% (see predicted trajectory in red). In this situation, the clinician is immediately alerted to the sudden increase in risk of dementia, and may decide to advise more frequent checkups, or to administer a wider range of tests and biomarker measurements in the immediate term to better assess the overall risk in light of the recent downturn. In fact, as it turns out in this case, the patient is indeed diagnosed with Alzheimer's disease at $t=10$ years, shedding light on \textsc{Match}-Net's potential as an early warning and subject selection system.

\section*{Related work}

\begin{table}[H]
\caption{Summary of primary improvements by related work\medskip}
\label{related_work}
\centering
\begin{tabular}{lcccccc} \toprule
Model & \thead{} & \thead{Non- \\ Linearity} & \thead{Deep \\ Learning} & \thead{Direct-to- \\ Probability} & \thead{Time- \\ Variance} & \thead{Dynamic \\ Prediction} \\ \midrule
Cox & (1972) & \textcolor{lightgray}{\xmark} & \textcolor{lightgray}{N/A} & \textcolor{lightgray}{N/A} & \textcolor{lightgray}{\xmark} & \textcolor{lightgray}{\xmark} \\
Faraggi \& Simon & (1995) & \cmark  & \textcolor{lightgray}{\xmark} & \textcolor{lightgray}{\xmark} & \textcolor{lightgray}{\xmark}  & \textcolor{lightgray}{\xmark} \\
Katzman et al. & (2016) & \cmark  & \cmark & \textcolor{lightgray}{\xmark} & \textcolor{lightgray}{\xmark} & \textcolor{lightgray}{\xmark} \\
Luck et al. & (2017) & \cmark & \cmark & \cmark & \textcolor{lightgray}{\xmark} & \textcolor{lightgray}{\xmark} \\
Lee et al. & (2018) & \cmark  & \cmark & \cmark & \cmark & \textcolor{lightgray}{\xmark} \\
\textbf{(This study)} & (2018) & \cmark  & \cmark & \cmark & \cmark & \cmark \\
\bottomrule
\end{tabular}
\end{table}

The first study to investigate neural networks formally in the context of time-to-event analysis was done by \cite{faraggi1995neural}. By swapping out the linear functional in the Cox model for the topology of a hidden layer, their nonlinear proportional hazards approach was extended to other models for censored data, such as \cite{rodriguez2005parametric} and \cite{buckley1979linear}. In 2016, \cite{katzman2016deep} were the first to apply modern techniques in deep learning to survival, in particular without prior feature selection or domain expertise. While previous studies following \cite{faraggi1995neural}'s model generally produced mixed results \cite{xiang2000comparison, mariani1997prognostic}, \cite{katzman2016deep} demonstrated comparable or superior performance of multilayer perceptrons in relation to conventional statistical methods.

Instead of predicting the hazard function as an intermediate objective, \cite{liestbl1994survival} first proposed\textemdash and \cite{biganzoli2009partial} further developed\textemdash an alternative approach to predict survival directly for grouped time intervals. In 2017, \cite{luck2017deep} combined the use of the Cox partial likelihood with the goal of predicting probabilities for pre-specified time intervals. Inspired by the work of \cite{yu2011learning} on multi-task logistic regression models for survival, they generalized the idea to deep learning via multi-task multilayer perceptrons.

Recently, \cite{lee2018deephit} and \cite{fotso2018deep} in 2018 proposed learning the distribution of survival times directly, making no assumptions regarding the underlying stochastic processes\textemdash in particular with respect to the time-invariance of hazards. By being process-agnostic, \cite{lee2018deephit} demonstrated significant improvements over existing statistical, machine learning, and neural network survival models on multiple real and synthetic datasets. In the context of Alzheimer's disease, \cite{parisot2018disease} studied the use of medical image sequences for classifying patient progression. More pertinently, \cite{lim2018forecasting} used recurrent networks to forecast disease trajectories with ADNI data, but relied on the explicit assumption of exponential distributions. Finally, building on these developments, one of the main contributions of this study is the use of temporal convolutions for dynamic survival prediction\textemdash while making no assumptions, and allowing the associations between covariates and risks to evolve over time (see Table \ref{related_work}).

\section*{Details on dataset}

We are primarily interested in the clinical status for each patient at any given time. An official diagnosis is recorded at each patient's visit, and consists of two attributes. First, each diagnosis may be either stable or transitive. The former consists of stable diagnoses of normal brain functioning (“NL”), mild cognitive impairment (“MCI”), or Alzheimer's disease (“AD”), and the latter consists of preliminary diagnoses indicating transitions between these categories, which may take the form of either conversions or reversions. Conversions indicate a forward progression in the disease trajectory, and reversions indicate a regression back towards an earlier stage of the disease.

\begin{figure}[H]
\centering
\includegraphics[width=1.0\linewidth]{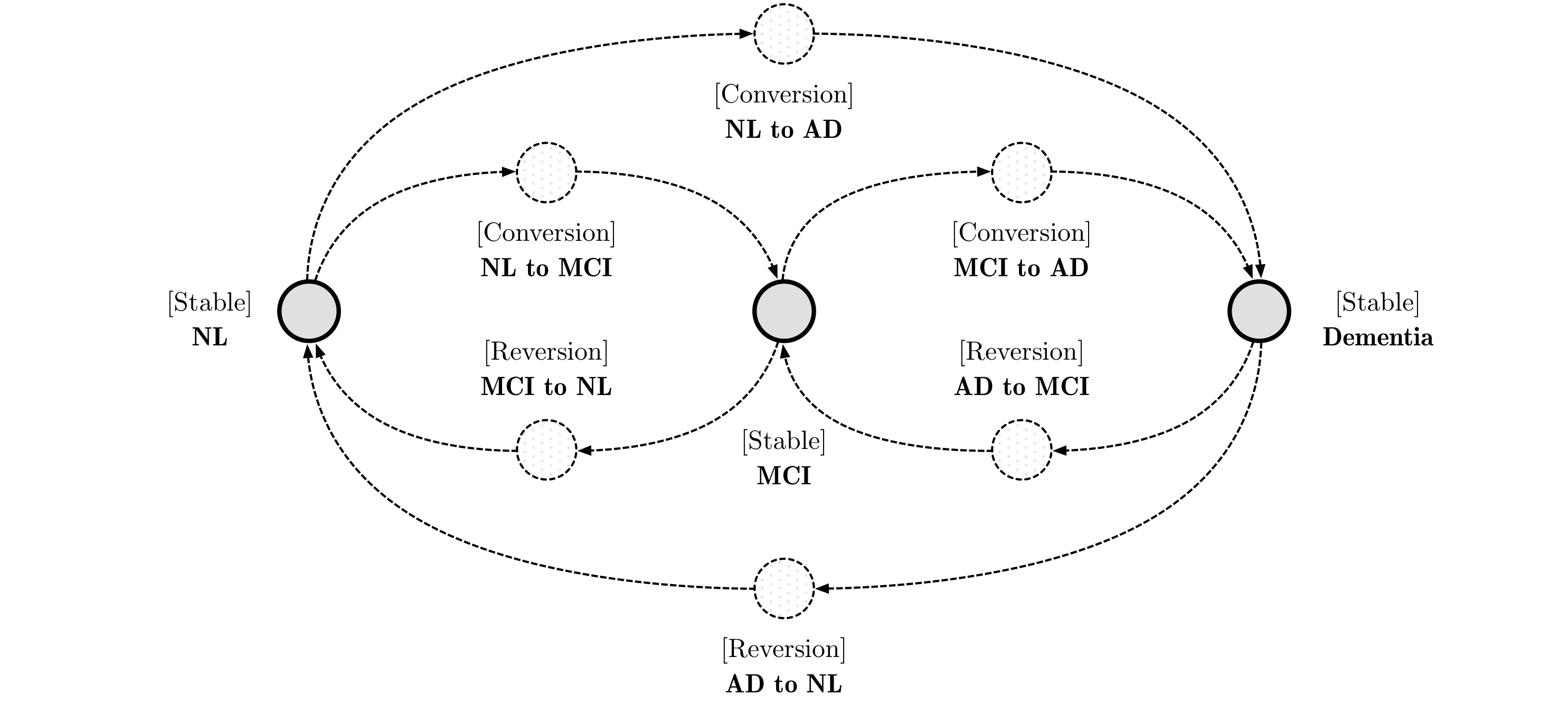}
\caption{State space of clinical diagnoses.}
\label{state_space}
\end{figure}

Patients may remain in stable or transition diagnosis states for any duration at a time. The average patient who receives a transition diagnosis is observed to persist in that state for one year, while some patients do not exit this state until almost 5 years have elapsed. Patients who receive a transition diagnosis may not actually be confirmed with a subsequent stable diagnosis; in fact, less than half of the transition diagnoses for dementia were confirmed by a stable diagnosis at the next step, and almost one quarter are never followed by a stable diagnosis at any point until right-censoring. In addition, patients often actually undergo reversion transitions back towards earlier stages of the disease; in fact, over 5\% of the study population receive reversion diagnoses at some point in time. 

\begin{table}
\caption{Summary and description of variables used in ADNI dataset.\smallskip}
\label{data_description}
\centering
\begin{tabular}{lllrrrrr} \toprule
{} & {} & {Type} & {Min} & {Max} & {Mean} & {S.D.} & {Missing} \\ \midrule
\multicolumn{2}{l}{\textbf{Event} (AD)} & Categorical & - & - & - & - & 30.1\% \\ \cmidrule{1-8}
\multicolumn{3}{l}{\textbf{Static}} \\
& Age & Numeric & 5.4E+01 & 9.1E+01 & 7.4E+01 & 7.2E+00 & 0.0\% \\
& APOE4 (Risk) & Numeric & 0.0E+00 & 2.0E+00 & 5.4E-01 & 6.6E-01 & 0.1\% \\
& Education Level & Numeric & 4.0E+00 & 2.0E+01 & 1.6E+01 & 2.9E+00 & 0.0\% \\
& Ethnicity & Categorical & - & - & - & - & 0.0\% \\
& Gender & Categorical & - & - & - & - & 0.0\% \\
& Marital Status & Categorical & - & - & - & - & 0.0\% \\
& Race & Categorical & - & - & - & - & 0.0\% \\ \cmidrule{1-8}
\multicolumn{3}{l}{\textbf{Biomarker}} \\
& Entorhinal & Numeric & 1.0E+03 & 6.7E+03 & 3.4E+03 & 8.1E+02 & 49.2\% \\
& Fusiform & Numeric & 7.7E+03 & 3.0E+04 & 1.7E+04 & 2.8E+03 & 49.2\% \\
& Hippocampus & Numeric & 2.2E+03 & 1.1E+04 & 6.7E+03 & 1.2E+03 & 46.6\% \\
& Intracranial & Numeric & 2.9E+02 & 2.1E+06 & 1.5E+06 & 1.7E+05 & 37.6\% \\
& Mid Temp & Numeric & 8.0E+03 & 3.2E+04 & 1.9E+04 & 3.1E+03 & 49.2\% \\
& Ventricles & Numeric & 5.7E+03 & 1.6E+05 & 4.2E+04 & 2.3E+04 & 41.6\% \\
& Whole Brain & Numeric & 6.5E+05 & 1.5E+06 & 1.0E+06 & 1.1E+05 & 39.7\% \\ \cmidrule{1-8}
\multicolumn{3}{l}{\textbf{Cognitive}} \\
& ADAS (11-item) & Numeric & 0.0E+00 & 7.0E+01 & 1.1E+01 & 8.6E+00 & 30.1\% \\
& ADAS (13-item) & Numeric & 0.0E+00 & 8.5E+01 & 1.8E+01 & 1.2E+01 & 30.7\% \\
& CRD Sum of Boxes & Numeric & 0.0E+00 & 1.8E+01 & 2.2E+00 & 2.8E+00 & 29.7\% \\
& Mini Mental State & Numeric & 0.0E+00 & 3.0E+01 & 2.7E+01 & 4.0E+00 & 29.9\% \\
& RAVLT Forgetting & Numeric & -1.2E+01 & 1.5E+01 & 4.2E+00 & 2.5E+00 & 30.9\% \\
& RAVLT Immediate & Numeric & 0.0E+00 & 7.5E+01 & 3.5E+01 & 1.4E+01 & 30.7\% \\
& RAVLT Learning & Numeric & -5.0E+00 & 1.4E+01 & 4.0E+00 & 2.8E+00 & 30.7\% \\
& RAVLT Percent & Numeric & -5.0E+02 & 1.0E+02 & 6.0E+01 & 3.8E+01 & 31.4\% \\
\bottomrule
\end{tabular}
\end{table}

\textbf{Note on class imbalance}. The per-patient failure rate is 14\% (243 patients out of the total 1,737). However, given the online nature of the sliding window mechanism in training and testing, the effective fraction of observations with positive event labels for any prediction horizon is around 2\%.

\section*{Data preparation}

Since the ADNI dataset is an amalgamation from multiple related studies, most features are sparsely populated. Features with less than half of the entries missing are retained, leaving 18 numeric and 4 categorical features (see Table \ref{data_description}); the latter are represented by one-hot encoding, resulting in 16 binary features. Consistent with existing Alzheimer's studies, patients are aligned according to time elapsed since baseline measurements \cite{li2017functional, liu2017prediction, vos2015prevalence}. Timestamps are mapped onto a discrete axis with a fixed resolution of $\delta=$ \nicefrac{1}{2}-year intervals; where multiple measurements qualify for the same destination, the most recent measurement per feature takes precedence. Original measurements were made at roughly \nicefrac{1}{2}-year intervals, so we observe that the average absolute deviation between original values and final timestamps amounts to an insignificant 4 days (\textit{i.e.} less than 2\% of each interval).

Where measurements are missing, values are reconstructed using zero-order hold interpolation. In addition, due to the fixed-width nature of the sliding window, the input tensor $\mathbf{X}_{i,t,w}$ for initial prediction times $t<w-\delta$ correspond to left-truncated information $t-w+\delta<0$; feature values are therefore extrapolated backwards for all intervals of the form $[-w+\delta,-\delta]$. Note that regardless of the imputation mechanism, information on original patterns of missingness\textemdash due to truncation, irregular sampling, and asynchronous sampling alike \cite{rubin1976inference, bangphased}\textemdash is preserved in the missing-value mask $\mathbf{Z}_{i,t,w}$ provided in parallel to the network. Finally, to improve numerical conditioning, all features are normalized with empirical means and standard deviations from the training set data.

\section*{Training procedure}

Training begins with input tuple $(\textsf{\textbf{X}},\textsf{\textbf{Z}},\mathbf{R})$, where $\textsf{\textbf{X}}=\{\langle\mathbf{X}_{i,t,w}\rangle^{t_i}_{t=0}\}^N_{i=1}$, $\textsf{\textbf{Z}}=\{\langle\mathbf{Z}_{i,t,w}\rangle^{t_i}_{t=0}\}^N_{i=1}$, and $\mathbf{R}=\{\langle\mathbf{r}_{i,t}\rangle^{t_i}_{t=0}\}^N_{i=1}$, and terminates with a set of calibrated network weights $\bm{\theta}$. The network is trained until convergence, up to a maximum of 50 epochs. Training loss is only computed for event labels corresponding to actual recorded clinical visits (\textit{i.e.} timestamps with recorded covariate values); neither imputed nor forward-filled labels are included. Analogous to our total loss function definition, the convergence metric is the sum of performance scores across all prediction tasks,
\vspace{-1.5mm} %%%
\begin{equation}
\label{convergence}
\mathcal{C}=\sum_{k=1}^{\tau_\text{max}/\delta}\beta(k\delta)\cdot\text{AUROC}_{k\delta}+\gamma(k\delta)\cdot\text{AUPRC}_{k\delta}
\vspace{-1.5mm} %%%
\end{equation}
where $\beta(\tau)$, $\gamma(\tau)$ optionally allow trading off the relative importance between the two measures and different horizons. In this study we simply use the unweighted sum, although any convex combination would be valid. Empirically, results are not meaningfully improved by favoring one metric over the other. Elastic net regularization is used, and validation performance is computed every 10 iterations. For early stopping, validation scores serve as proxies for generalization error. Positive instances are oversampled to counteract class imbalance. To augment training data, artificial labels for dummy events of the form $\tau=0$ are generated, and positive labels are filled forward for all horizons $\tau>\tau_0$ where $\tau_0$ corresponds to the first failure for that example. Patients diagnosed with Alzheimer's at baseline (20\%) are excluded from testing and validation, since survival is undefined in those cases.

\begin{table}[H]
\caption{Hyperparameter selection ranges for random search.\smallskip}
\label{hp_optimization}
\centering
\begin{tabular}{ll} \toprule
{Hyperparameter} & {Selection Range} \\ \midrule
Connected Layers & 1, 2, 3, 4, 5\\
Convolutional Layers & 1, 2, 3, 4, 5\\
Dropout Rate & 0.1, 0.2, 0.3, 0.4, 0.5\\
Epochs for Convergence & 1, 2, 3, 4, 5, 6, 7, 8, 9, 10\\
Learning Rate & 1e-4, 3e-4, 1e-3, 3e-3, 1e-2, 3e-2\\
L1-Regularlisation & None, 1e-4, 3e-4, 1e-3, 3e-3, 1e-2, 3e-2, 1e-1\\
L2-Regularization & None, 1e-4, 3e-4, 1e-3, 3e-3, 1e-2, 3e-2, 1e-1\\
Minibatch Size & 32, 64, 128, 256, 512\\
Number of Filters (Covariates) & 32, 64, 128, 256, 512\\
Number of Filters (Masks) & 8, 16, 32, 64, 128\\
Oversample Ratio & None, 1, 2, 3, 5, 10\\
Recurrent Unit State Size & 1$\times$, 2$\times$, 3$\times$, 4$\times$, 5$\times$\\
Width of Connected Layers & 32, 64, 128, 256, 512\\
Width of Convolutional Filters & 3, 4, 5, 6, 7, 8, 9, 10\\
Width of Sliding Window & 3, 4, 5, 6, 7, 8, 9, 10\\
\bottomrule
\end{tabular}
\end{table}

For Cox landmarking, we use the implementation in \cite{terry2018package} for interval-censored data, fitting a sequence of proportional hazards regression models for observation groups. Optimal groupings are determined by exhaustive search in \nicefrac{1}{2}-year increments. Preliminary feature-selection is performed by stepwise regression for the best candidate model, using the implementation of \cite{hu2018package}. For joint modeling, we adopt the common two-stage method in \cite{wu2009mixed}, first fitting linear mixed effects sub-models for significant variables, then fitting landmarking models based on mean estimates from the sub-models. In both cases, consistent with literature, time is defined as years since initial follow-up \cite{li2017functional, liu2017prediction, vos2015prevalence}.

For all neural network models, hyperparameter optimization is carried out via 100 iterations of random search(see Table \ref{hp_optimization}). In addition, activation functions are also searched over for gating units in recurrent network cells. Model selection is performed on the basis of final composite scores\textemdash as in Equation \ref{convergence}\textemdash for each candidate. We use 5-fold cross validation to evaluate performance, stratified at the patient level\textemdash that is, patients are randomly selected into datasets for training (60\%), validation (20\%), and testing (20\%), with the ratio of positive patients (\textit{i.e.} those for which at least one sliding window contains at least one outcome of failure) to negative patients is kept uniform across folds.

\section*{Further analysis}

\begin{figure}[H]
\centering
\includegraphics[width=0.97\linewidth]{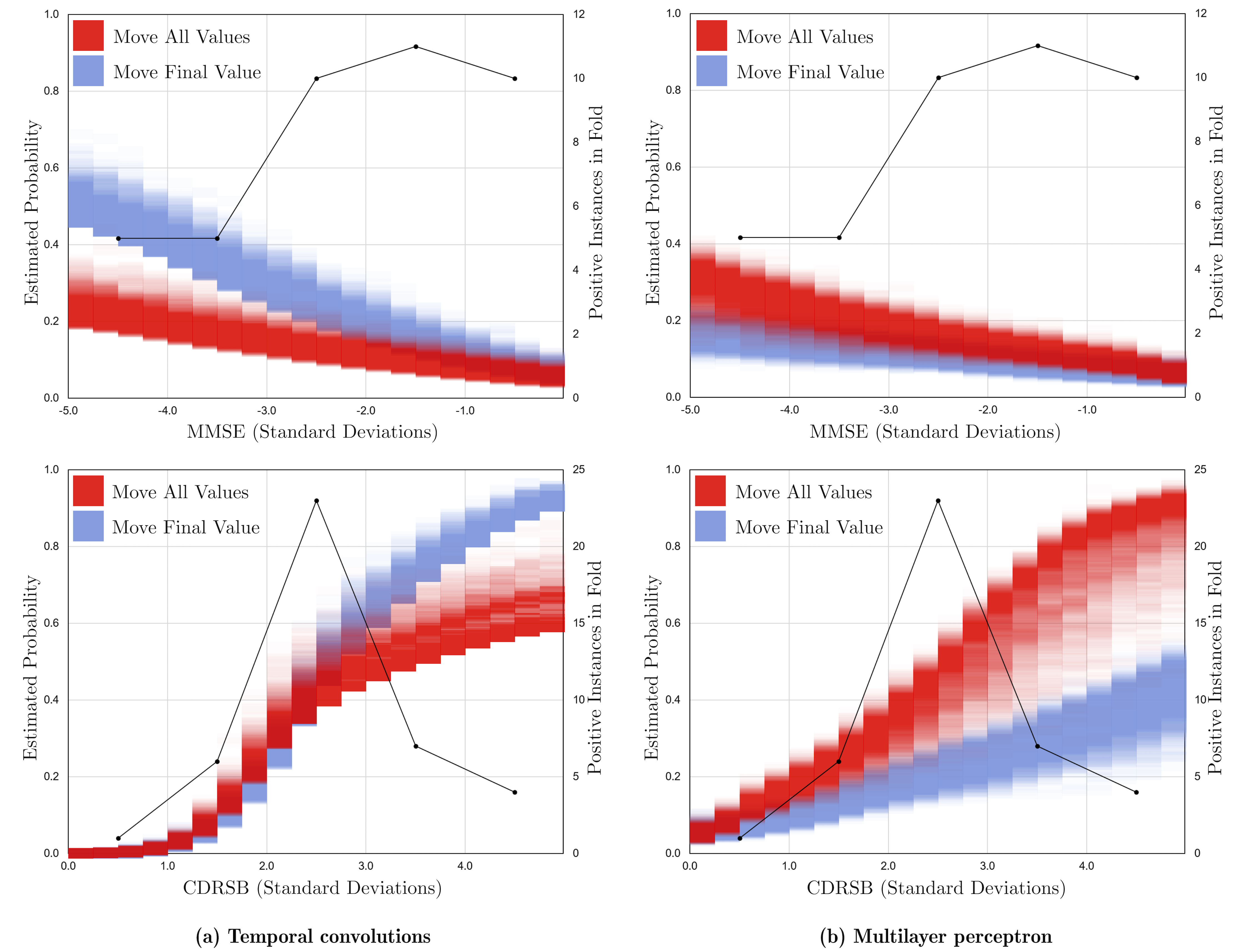}
\caption{Comparison of output scatters for changes in all values within sliding window (red) versus changes in the final value only (blue). Convolutional models show stronger responses in the latter.}
\label{output_scatter}
\end{figure}

In addition, we utilize \textit{output scatters} to aid our understanding of the input-output relationship. We use MC dropout to generate average responses by explicitly varying one input feature at a time \cite{gal2016dropout} in equation \ref{partial_dependence}. This is done by either (1) varying only the final value of the measurement within the sliding window, or by (2) varying all values of the measurement the same time. While the latter gives insight into how the response changes with respect to a difference in \textit{levels}, the former sheds light on how the response is affected by a sudden \textit{change} instead. Interestingly in the case of temporal convolutions, for multiple features (Figure \ref{output_scatter} shows the MMSE and CDRSB features as examples) the response is actually \textit{stronger} to final value changes. This effect is not present in any feature for multilayer perceptrons. In light of the performance of convolutional models, this is consistent with our hypothesis that they appear able to learn more complex temporal patterns than multilayer perceptrons; in this case, these results suggest that a patient who experiences a sudden decline in certain features is expected to fare worse than one who has always scored just as poorly all along.

\section*{Sources of gain}

While the advantage of using multilayer perceptrons over traditional statistical survival models has been studied (see \cite{lee2018deephit} for example), here we account for the additional sources of gain from our design choices. Table \ref{sources_of_gain}~(a) shows the initial benefit from incorporating \textit{longitudinal histories} of covariate measurements. Recurrent networks are used as a reasonable starting point; improvements\textemdash where positive\textemdash are marginal at best. (b) However, experiments on differenced data\textemdash capturing relative movements without absolute levels\textemdash indicate substantial informativeness (see \textit{df}-RNN trained on \textit{differences} vs. joint models trained on \textit{levels}). (c) Utilizing limited \textit{sliding windows} instead of recurrent cells shows promising improvements, boosting average AUPRC by 4\% over both RNN and MLP models. (d) The addition of \textit{temporal convolutions} produces incremental AUPRC gains of 9\%. (e) Accommodating \textit{informative missingness} by introducing the dual-stream architecture results in further incremental AUPRC gains of 2\%. Compared with the  joint modeling baseline, \textsc{Match}-Net achieves average AUPRC improvements of 15\% and one-step-ahead improvements of 26\%. (f) Furthermore, incorporating the most recent clinician input (\textsc{Match}-Net$^{+}$) improves AUPRC by an additional 17\%. While the distribution of gains skews near-term with our default choices of $\alpha$, $\beta$, and $\gamma$ (\textit{i.e.} unweighted by $\tau$), alternate distributions may be obtained via appropriate weight functions depending on deployment context. (g) Finally, using \textsc{Match}-Net as an example, we observe individual and cumulative benefits due to miscellaneous design choices applied to all models. 

\begin{table}
\caption{Source-of-gain accounting. Bold values indicate best performance. Note that AUROC is much less sensitive than AUPRC in the context of highly imbalanced classes like the ADNI data \cite{saito2015precision}.\medskip}
\label{sources_of_gain}
\centering
\subfloat[Gain from Covariate History]{
\begin{tabular}{c|c|cccc} \toprule
{} & {$\tau$} & \multicolumn{2}{c}{RNN} & \multicolumn{2}{c}{MLP} \\ \cmidrule{1-6}
\textsc{auroc} & 0.5  & \textbf{0.949} & \textbf{$\pm\text{0.009}$} & 0.948 & $\pm\text{0.010}$ \\
& 1.0  & \textbf{0.930} & \textbf{$\pm\text{0.012}$} & \textbf{0.930} & \textbf{$\pm\text{0.011}$} \\
& 1.5  & \textbf{0.891} & \textbf{$\pm\text{0.026}$} & 0.890 & $\pm\text{0.027}$ \\
& 2.0  & \textbf{0.901} & \textbf{$\pm\text{0.025}$} & 0.895 & $\pm\text{0.029}$ \\
& 2.5  & \textbf{0.883} & \textbf{$\pm\text{0.031}$} & 0.874 & $\pm\text{0.039}$ \\ \cmidrule{1-6}
\textsc{auprc} & 0.5  & 0.464 & $\pm\text{0.079}$ & \textbf{0.469} & \textbf{$\pm\text{0.064}$} \\
& 1.0  & 0.410 & $\pm\text{0.060}$ & \textbf{0.435} & \textbf{$\pm\text{0.056}$} \\
& 1.5  & \textbf{0.340} & \textbf{$\pm\text{0.067}$} & \textbf{0.340} & \textbf{$\pm\text{0.067}$} \\
& 2.0  & 0.355 & $\pm\text{0.068}$ & \textbf{0.359} & \textbf{$\pm\text{0.065}$} \\
& 2.5  & \textbf{0.365} & \textbf{$\pm\text{0.087}$} & 0.356 & $\pm\text{0.085}$ \\
\bottomrule
\end{tabular}
}
\subfloat[Informativeness of Differences]{
\begin{tabular}{cccccc} \toprule
{} & \multicolumn{2}{c}{\textit{df}-RNN} & \multicolumn{2}{c}{JM} \\ \cmidrule{2-5}
& \textbf{0.913} & \textbf{$\pm\text{0.026}$} & \textbf{0.913} & \textbf{$\pm\text{0.036}$} \\
& 0.830 & $\pm\text{0.020}$ & \textbf{0.917} & \textbf{$\pm\text{0.016}$} \\
& 0.700 & $\pm\text{0.057}$ & \textbf{0.881} & \textbf{$\pm\text{0.029}$} \\
& 0.712 & $\pm\text{0.059}$ & \textbf{0.894} & \textbf{$\pm\text{0.034}$} \\
& 0.708 & $\pm\text{0.083}$ & \textbf{0.883} & \textbf{$\pm\text{0.036}$} \\ \cmidrule{2-5}
& \textbf{0.494} & \textbf{$\pm\text{0.057}$} & 0.473 & $\pm\text{0.072}$ \\
& 0.386 & $\pm\text{0.036}$ & \textbf{0.415} & \textbf{$\pm\text{0.042}$} \\
& 0.213 & $\pm\text{0.058}$ & \textbf{0.319} & \textbf{$\pm\text{0.044}$} \\
& 0.241 & $\pm\text{0.078}$ & \textbf{0.362} & \textbf{$\pm\text{0.050}$} \\
& 0.217 & $\pm\text{0.096}$ & \textbf{0.366} & \textbf{$\pm\text{0.064}$} \\
\bottomrule
\end{tabular}
}

\subfloat[Gain from Limited Window]{
\begin{tabular}{c|c|cccc} \toprule
{} & {$\tau$} & \multicolumn{2}{c}{S-MLP} & \multicolumn{2}{c}{RNN} \\ \cmidrule{1-6}
\textsc{auroc} & 0.5  & \textbf{0.950} & \textbf{$\pm\text{0.009}$} & 0.949 & $\pm\text{0.009}$ \\
& 1.0  & \textbf{0.932} & \textbf{$\pm\text{0.012}$} & 0.930 & $\pm\text{0.012}$ \\
& 1.5  & \textbf{0.897} & \textbf{$\pm\text{0.025}$} & 0.891 & $\pm\text{0.026}$ \\
& 2.0  & \textbf{0.904} & \textbf{$\pm\text{0.026}$} & 0.901 & $\pm\text{0.025}$ \\
& 2.5  & 0.881 & $\pm\text{0.035}$ & \textbf{0.883} & \textbf{$\pm\text{0.031}$} \\ \cmidrule{1-6}
\textsc{auprc} & 0.5  & \textbf{0.500} & \textbf{$\pm\text{0.066}$} & 0.464 & $\pm\text{0.079}$ \\
& 1.0  & \textbf{0.447} & \textbf{$\pm\text{0.056}$} & 0.410 & $\pm\text{0.060}$ \\
& 1.5  & \textbf{0.354} & \textbf{$\pm\text{0.061}$} & 0.340 & $\pm\text{0.067}$ \\
& 2.0  & \textbf{0.364} & \textbf{$\pm\text{0.054}$} & 0.355 & $\pm\text{0.068}$ \\
& 2.5  & \textbf{0.371} & \textbf{$\pm\text{0.084}$} & 0.365 & $\pm\text{0.087}$ \\
\bottomrule
\end{tabular}
}
\subfloat[Gain from Temporal Convolutions]{
\begin{tabular}{cccccc} \toprule
{} & \multicolumn{2}{c}{S-TCN} & \multicolumn{2}{c}{S-MLP} \\ \cmidrule{2-5}
& \textbf{0.961} & \textbf{$\pm\text{0.005}$} & 0.950 & $\pm\text{0.009}$ \\
& \textbf{0.941} & \textbf{$\pm\text{0.007}$} & 0.932 & $\pm\text{0.012}$ \\
& \textbf{0.902} & \textbf{$\pm\text{0.025}$} & 0.897 & $\pm\text{0.025}$ \\
& \textbf{0.908} & \textbf{$\pm\text{0.026}$} & 0.904 & $\pm\text{0.026}$ \\
& \textbf{0.884} & \textbf{$\pm\text{0.032}$} & 0.881 & $\pm\text{0.035}$ \\ \cmidrule{2-5}
& \textbf{0.580} & \textbf{$\pm\text{0.066}$} & 0.500 & $\pm\text{0.066}$ \\
& \textbf{0.505} & \textbf{$\pm\text{0.065}$} & 0.447 & $\pm\text{0.056}$ \\
& \textbf{0.367} & \textbf{$\pm\text{0.063}$} & 0.354 & $\pm\text{0.061}$ \\
& \textbf{0.380} & \textbf{$\pm\text{0.052}$} & 0.364 & $\pm\text{0.054}$ \\
& \textbf{0.381} & \textbf{$\pm\text{0.085}$} & 0.371 & $\pm\text{0.084}$ \\
\bottomrule
\end{tabular}
}

\subfloat[Gain from Missingness-Awareness]{
\begin{tabular}{c|c|cccc} \toprule
{} & {$\tau$} & \multicolumn{2}{c}{\textsc{Match}-Net} & \multicolumn{2}{c}{S-TCN} \\ \cmidrule{1-6}
\textsc{auroc} & 0.5  & \textbf{0.962} & \textbf{$\pm\text{0.004}$} & 0.961 & $\pm\text{0.005}$ \\
& 1.0  & \textbf{0.942} & \textbf{$\pm\text{0.007}$} & 0.941 & $\pm\text{0.007}$ \\
& 1.5  & \textbf{0.902} & \textbf{$\pm\text{0.024}$} & \textbf{0.902} & \textbf{$\pm\text{0.025}$} \\
& 2.0  & \textbf{0.909} & \textbf{$\pm\text{0.027}$} & 0.908 & $\pm\text{0.026}$ \\
& 2.5  & \textbf{0.886} & \textbf{$\pm\text{0.033}$} & 0.884 & $\pm\text{0.032}$ \\ \cmidrule{1-6}
\textsc{auprc} & 0.5  & \textbf{0.594} & \textbf{$\pm\text{0.058}$} & 0.580 & $\pm\text{0.066}$ \\
& 1.0  & \textbf{0.513} & \textbf{$\pm\text{0.059}$} & 0.505 & $\pm\text{0.065}$ \\
& 1.5  & \textbf{0.373} & \textbf{$\pm\text{0.065}$} & 0.367 & $\pm\text{0.063}$ \\
& 2.0  & \textbf{0.390} & \textbf{$\pm\text{0.059}$} & 0.380 & $\pm\text{0.052}$ \\
& 2.5  & \textbf{0.384} & \textbf{$\pm\text{0.081}$} & 0.381 & $\pm\text{0.085}$ \\
\bottomrule
\end{tabular}
}
\subfloat[Gain from Most Recent Diagnosis]{
\begin{tabular}{cccccc} \toprule
{} & \multicolumn{2}{c}{\textsc{Match}-Net$^{+}$} & \multicolumn{2}{c}{\textsc{Match}-Net} \\ \cmidrule{2-5}
& \textbf{0.989} & \textbf{$\pm\text{0.003}$} & 0.962 & $\pm\text{0.004}$ \\
& \textbf{0.951} & \textbf{$\pm\text{0.008}$} & 0.942 & $\pm\text{0.007}$ \\
& 0.897 & $\pm\text{0.026}$ & \textbf{0.902} & \textbf{$\pm\text{0.024}$} \\
& 0.901 & $\pm\text{0.024}$ & \textbf{0.909} & \textbf{$\pm\text{0.027}$} \\
& 0.885 & $\pm\text{0.024}$ & \textbf{0.886} & \textbf{$\pm\text{0.033}$} \\ \cmidrule{2-5}
& \textbf{0.862} & \textbf{$\pm\text{0.037}$} & 0.594 & $\pm\text{0.058}$ \\
& \textbf{0.636} & \textbf{$\pm\text{0.043}$} & 0.513 & $\pm\text{0.059}$ \\
& \textbf{0.393} & \textbf{$\pm\text{0.082}$} & 0.373 & $\pm\text{0.065}$ \\
& 0.380 & $\pm\text{0.051}$ & \textbf{0.390} & \textbf{$\pm\text{0.059}$} \\
& 0.375 & $\pm\text{0.066}$ & \textbf{0.384} & \textbf{$\pm\text{0.081}$} \\
\bottomrule
\end{tabular}
}

\subfloat[Gain from oversampling (O), label forwarding (L), and elastic net (E) for \textsc{Match}-Net]{
\begin{tabular}{c|ccccc|ccccc} \toprule
\multicolumn{1}{c}{} & \multicolumn{5}{c}{AUROC} & \multicolumn{5}{c}{AUPRC} \\ \cmidrule(l{2pt}r{2pt}){2-6} \cmidrule(l{2pt}r{2pt}){7-11}
\multicolumn{1}{c}{O~~L~~E} & {0.5} & {1.0} & {1.5} & {2.0} & \multicolumn{1}{c}{2.5} & {0.5} & {1.0} & {1.5} & {2.0} & {2.5} \\ \midrule
\textcolor{lightgray}{\xmark}~~\textcolor{lightgray}{\xmark}~~\textcolor{lightgray}{\xmark} & 0.946 & 0.923 & 0.895 & 0.902 & \textbf{0.888} & 0.493 & 0.418 & 0.348 & 0.351 & 0.355 \\
\textcolor{lightgray}{\xmark}~~\textcolor{lightgray}{\xmark}~~\cmark & 0.956 & 0.938 & 0.901 & 0.906 & 0.886 & 0.534 & 0.477 & 0.368 & 0.381 & \textbf{0.384} \\
\textcolor{lightgray}{\xmark}~~\cmark~~\textcolor{lightgray}{\xmark} & 0.944 & 0.920 & 0.888 & 0.896 & 0.886 & 0.435 & 0.413 & 0.334 & 0.351 & 0.357 \\
\textcolor{lightgray}{\xmark}~~\cmark~~\cmark & 0.961 & 0.941 & \textbf{0.902} & 0.908 & 0.887 & 0.575 & 0.502 & 0.370 & 0.384 & 0.377 \\
\cmark~~\textcolor{lightgray}{\xmark}~~\textcolor{lightgray}{\xmark} & 0.946 & 0.923 & 0.894 & 0.902 & 0.887 & 0.499 & 0.425 & 0.352 & 0.365 & 0.365 \\
\cmark~~\textcolor{lightgray}{\xmark}~~\cmark & 0.956 & 0.938 & 0.901 & 0.907 & 0.887 & 0.533 & 0.473 & 0.362 & 0.379 & 0.385 \\
\cmark~~\cmark~~\textcolor{lightgray}{\xmark} & 0.942 & 0.918 & 0.885 & 0.894 & 0.881 & 0.448 & 0.404 & 0.332 & 0.348 & 0.361 \\
\cmark~~\cmark~~\cmark & \textbf{0.962} & \textbf{0.942} & \textbf{0.902} & \textbf{0.909} & 0.886 & \textbf{0.594} & \textbf{0.513} & \textbf{0.373} & \textbf{0.390} & \textbf{0.384} \\
\bottomrule
\end{tabular}
}
\end{table}

\end{document}